# Structure and Motion from Multiframes

Mieczysław A. Kłopotek

*Institute of Computer Science PAS, 01-237 Warsaw, ul. Ordona 21,Poland, klopotek@ipipan.waw.pl*

**Abstract**

The paper gives an overview of the problems and methods of recovery of structure and motion parameters of rigid bodies from multiframes.

## 1 Introduction

The evolution of animal optical systems (eyes) led to elimination of static elements from the seen scene, leaving only moving parts of the picture for further processing by the central neuro-system. It was of importance for survival because food and enemy were usually moving objects.

By analogy, it may be assumed that for a number of vision systems moving objects will be also of primary interest (robots at production lines, collision avoiding systems etc.)

Shape (structure) reconstruction of 3D space objects from 2D images [12, 13, 14, 15, 17, 4, 21, 31, 5, 27] is heavily under-constrained, unless shape [54, 55, 2, 41], texture [42], shadows [52] or other clues are available. As a way out of this problem, 3-D sensory methods are frequently applied, based on sensing (via laser beams, ultrasonic methods etc. [40, 35], or other active vision techniques. The 3-D sensory methods proved useful when reconstructing voluminous objects. They are, however, not quite well suited for outspoken smooth curve-like object.

So this remains still a research area for 2D projection based recognition methods. Some promising results were in fact achieved in recovering objects from multiframes (a time sequence of projections of the moving object) [50, 24, 3, 28, 51, 7, 8, 36, 22, 1, 45]. as this task is over-constrained. Also in cases where features of interest cannot be all traced from frame to frame - e.g. smooth-curve shaped objects [38, 29, 33] In fact, only several features (usually end points) are traceable, and the remaining ones are not.

Though we are interested here only in objects consisting of 0 and 1-dimentional elements (points, lines, curves), polyhedral objects are also investigated intensely in the structure-from-motion area [16, 32, 37].

An essential assumption is the rigidness of the body, though in some cases it may be weakened [61, 22]

Though structure recovering while motion is known is a challenging task, [43, 59, 62, 49], most researchers assume that both motion and structure are unknown.

The task of identifying structure and motion of 3-D objects may be roughly divided into three parts:

- identification of the moving object in the scene

- Matching object features (points, lines etc.) between the images - establishing so-called traced points, traced lines etc.

- Proper reconstruction of structure and motion using the detected traced features, which may be further refined into

    - reconstruction of the structure of traced feature elements
    - matching and reconstruction of the structure of non-feature (not traced) elements.

Though the first two stages are surely very complex and important from the technical point of view [44, 11, 23], the last stage has been paid much attention in research effort. This paper will concentrate especially on this third stage of the whole problem.

The research in the area concentrated in general on the following topics:

- to discover methods of reconstruction of spatial position of traceable features (points, line segments) from multiframes

- to find the minimum number of traceable features needed or the minimum number of frames needed

- to find tractable algorithms of reconstruction (linear equation systems, special kinds of motion)

- to elaborate methods increasing reliability of recovered structures (usually via increasing the number of frames)

The problem has been considered under the following geometrical constraints:

- orthogonal projections

- perspective projections if the relative position of focal point with respect to the projection plane is known and

- perspective projections if the relative position of focal point with respect to the projection plane is not known ("uncalibrated cameras").

The paper will concentrate on the minimal number of features (lines, points, parallel line beams) needed for structure and motion recovery as well as on the trade-off between the number of features, frames and solution complexity.

The case of orthogonal projections may be perceived and in fact is a significant simplification of the reconstruction task. As few as two feature (traced) points are needed if uniform rotation around a fixed direction is assumed. However, it turns out that under orthogonal projection two frames are insufficient for recovery of structure and motion.

Though under perspective projections this is in general no longer the case, the complexity of equations to be solved is so high that iterative methods are needed providing solutions even in cases where none exist, as we run at risk of overseeing various geometrical constraints imposed on frames. This led to the situation that some published results later turned out to be false just for that reason.

## 2 Over/underconstraining - the problem of degrees of freedom

### 2.1 Degrees of freedom for orthogonal projection

Each point introduces 3 df in the first frame, each line - 4 df minus one df for the whole body as there exists no possibility of determining the initial depth of the body in the space. The motion introduces for each subsequent frame 5 df only, because the motion in the direction orthogonal to the projection plane has no impact on the image. In general, with p points and s straight lines forming the rigid body traced over k frames we have

$$-1 + 3*p + 4*s + 5*(k-1)$$

degrees of freedom against

$$k*(2*p + 2*s)$$

pieces of information available from k images.
Thus we shall have the balance

$$-1 + 3*p + 4*s + 5*(k-1) \leq k*(2*p + 2*s) \quad (1)$$

to achieve recoverability.

Let us consider some combinations of parameters:

- for k=3 frames, p=3 points we get

$$-1 + 3*p + 4*s + 5*(k-1) = 18 = k*(2*p + 2*s) = 18$$

- for k=2 frames, p=4 points we get

$$-1+3*p+4*s+5*(k-1) = -1+12+5 = 16 = k*(2*p+2*s) = 2*2*4 = 16$$

On exploiting straight line component of the above equation see [57], and on non-geometrical balancing degrees of freedom see [43].

As the minimum number of features and/or frames needed we have to state that though the above df-consideration would indicate that two frames may be sufficient, in [63] it has been demonstrated that this is not the case. With 2 frames, none of points beyond the 3rd contributes any information to the location of the object in 3d, because we can assign any point in the first frame a straight line in the second frame. We can generally state that the minimum number of frames needed for recovery is 3, the minimal number of traceable points is also 3.

## 2.2 Perspective projections with fixed focal point

In general, with $p$ points and $s$ straight lines forming the rigid body traced over k frames we have the following number of degrees of freedom:

$$-1 + 3*p + 4*s + 6*(k-1)$$

The constituent -1 is due to the fact that the scaling of the object cannot be recovered under perspective projection. The factor 3 means the number of degrees of freedom for a point, 4 - for a straight line and 6 - for the motion between frames.

Now the amount of information gained within those k frames amounts to:

$$k*(2*p+2*s)$$

In order to recover the structural and motion data we request that:

$$-1 + 3*p + 4*s + 6*(k-1) \le k*(2*p+2*s) \qquad (2)$$

When we have to do with 2 frames (k=2) and 4 points (p=4, s=0) only, we obtain:.

$$-1+3*p+4*s+6*(k-1) = -1+12+6 = 17 > k*(2*p+2*s) = 2*2*4 = 16$$

which means that the problem is underconstrained.

Let us notice, however, that with 3 frames (k=3) and 4 points (p=4, s=0) we obtain

$$-1+3*p+4*s+6*(k-1) = -1+12+12 = 23 < k*(2*p+2*s) = 3*2*4 = 24$$

ensuring the existence of a solution (see [47]

Also with 2 frames (k=2) and 5 points (p=5, s=0) we obtain

$$-1+3*p+4*s+6*(k-1) = -1+15+6 = 20 = k*(2*p+2*s) = 2*2*5 = 20$$

ensuring the existence of a solution (see [18, 19, 20]).

Also with 3 frames (k=3) and 3 points and a single line (p=3, s=1) we obtain

$$-1+3*p+4*s+6*(k-1) = -1+9+4+12 = 24 = k*(2*p+2*s) = 3*(6+2) = 24$$

ensuring the existence of a solution.

With 3 frames (k=3) and six lines (p=0, s=6) we obtain

$$-1+3*p+4*s+6*(k-1) = -1+0+24+12 = 35 < k*(2*p+2*s) = 3*(0+12) = 36$$

ensuring the existence of a solution (compare [60]).

The minimal number of frames needed for recovery is 2, the minimal number of points is 4 (though not with 2 but with 3 frames).

## 2.3 Unknown and moving projection focal point

Let us now consider the degrees of freedom for the perspective projection if we assume that the relative position (in space) of the focal point with respect to the projection plane is not known and may vary over time.

Each point of the body introduces 3 df in the first frame minus one df for the whole body as there exists no possibility of determining the scaling of the whole body. Additionally we have 3df due to the uncertainty of the location of the focal point. The motion introduces for each subsequent frame 9 df (three for rotations and three for translation of the projection plane plus three for translation of the focal point). In general, with p points forming the rigid body traced over k frames we have then $-1+3*p+3+9*(k-1)$ degrees of freedom. On the other hand, within each image each traced point provides us with two pieces of information: its x and its y position within the frame. Hence we have at most $k*2*p$ pieces of information available from k images. Thus we need at least to have the balance

$$-1+3*p+3+9*(k-1) \leq k*2*p \quad (3)$$

to achieve recoverability. Let us consider some combinations of parameters:

- for k=2 frames, p= 10 points we get $-1+3*p+3+9*(k-1) = 41 > k*2*p = 40$

- for k=2 frames, p= 11 points we get $-1+3*p+3+9*(k-1) = 44 = k*2*p = 44$

- for k=2 frames, p=7 points we get $-1+3*p+3+9*(k-1) = 32 > k*2*p = 28$

- for k=3 frames, p=7 points we get $-1 + 3*p + 3 + 9*(k-1) = 41 < k*2*p = 42$

- for k=3 frames, p=6 points we get $-1 + 3*p + 3 + 9*(k-1) = 38 > k*2*p = 36$

- for k=4 frames, p=6 points we get $-1 + 3*p + 3 + 9*(k-1) = 47 < k*2*p = 48$

- for k=8 frames, p=5 points we get $-1 + 3*p + 3 + 9*(k-1) = 80 = k*2*p = 80$

The above (in)equalities tell us that to recover structure and motion from 5 traceable points, we would need 8 images (frames), with 7 traceable points we need 3 frames, and to recover from two frames we would need 11 points - if we take the balance of degrees of freedom and the amount of information. However, as shown in [64], with 2 frames, none of points beyond the 7th contributes any information to the location of the object in 3d, because we can assign any point in the first frame a straight line in the second frame.

We claim that also three frames are insufficient, because with 3 frames each pair of points, one stemming from the first frame, the other from the second, can be assigned exactly one point in the third frame. The proof for this statement may be achieved by constructing different spatial arrangements of same 3 frames corresponding to different 3-d objects.

If we have only p=4 traceable points, then we get the number of degrees of freedom equal to -1+3*4+9*(k-1)=9k+2, whereas the amount of information is equal to k*2*4=8k, which is always less then the number of degrees of freedom. This means that if we trace only four points, we can never recover structure and motion whatever number of frames is available.

## 3 Finding solutions

### 3.1 Orthogonal projections - structure and motion for 3 point correspondences

Let us briefly sketch the procedure of recovery of a three-point structure from multiframes.

Let $P$, $Q$, $R$ be the traced points of a rigid body, and $P_i$, $Q_i$, $R_i$ their respective projections within the $i^{th}$ frame (Fig.1). Let $a$, $b$, $c$, $a_i$, $b_i$, $c_i$ denote the lengths of straight line segments $PQ$, $QR$, $RP$, $P_iQ_i$, $Q_iR_i$, $R_iP_i$, respectively. Then for each frame one of the following relationships holds: Either: $\sqrt{a^2 - a_i^2} + \sqrt{b^2 - b_i^2} - \sqrt{c^2 - c_i^2} = 0$ or $\sqrt{a^2 - a_i^2} - \sqrt{b^2 - b_i^2} + \sqrt{c^2 - c_i^2} = 0$ or $-\sqrt{a^2 - a_i^2} + \sqrt{b^2 - b_i^2} + \sqrt{c^2 - c_i^2} = 0$ (which is easily seen from geometrical relationships, presented analytically and graphically by Kłopotek[57]). So

Figure 1: Three points and their orthogonal projections. Length of $\overline{QQ_i}$ equals $\overline{P_Q P_i}$.

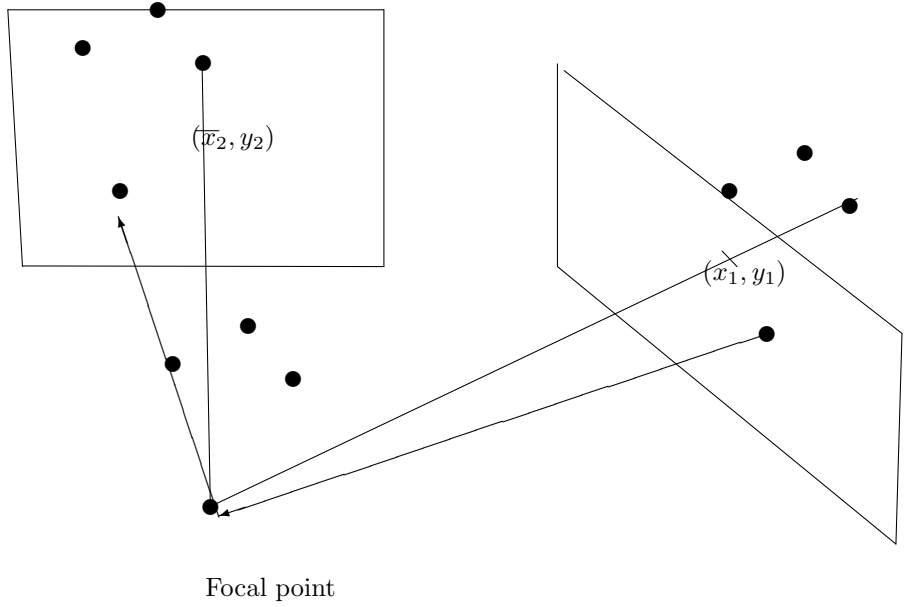

Focal point

Figure 2: Perspective projection

we have three equations, for $i = 1$, 2, and 3, in three unknowns, $a$, $b$, $c$. As any of the above relationships gives after a twofold squaring:

$$a^4 + b^4 + c^4 - 2a^2b^2 - 2a^2c^2 - 2b^2c^2 + a_i^4 + b_i^4 + c_i^4 - 2a_i^2b_i^2 - 2a_i^2c_i^2 - 2b_i^2c_i^2$$
$$+ 2(-a_i^2 + b_i^2 + c_i^2)a^2 + 2(+a_i^2 - b_i^2 + c_i^2)b^2 + 2(+a_i^2 + b_i^2 - c_i^2)c^2 = 0 \quad (4)$$

which is quadratic in $a^2$, $b^2$, $c^2$, hence solvable by exploitation of proper methods.

We used a partial linearization approach in that from formulas for $i = 1$ and $i = 2$ we subtracted with one for $i = 3$:

## 3.2 Perspective projections with fixed focal point

The basic approach consists in matching two subsequent scenes and considering the rotation matrix and translation vectors expressing motion from frame to frame [7] (Fig. 2). Assume a coordinate system XYZ with center point at the center of coordinate system at the camera focal point. Assume that one point of a rigid object consisting of n traceable points, lying on the Z axis (X,Y axes parallel to the projection plane) is 1 unit away from the focal point of the camera away in the first frame (Assumption is legal due to the scaling factor not recovered under perspective projection). We can "rotate" the second projection plane in such a way that this distinguished point again lies on the Z - axis (done by simple recalculation). For simplicity, assume that the coordinate system center in the projections plane lies on the universal Z axis one unit away from the focal point and the xy coordinate lines of the projection plane are parallel to the XY axes. Then we can assume that the whole object has moved to place the distinguished point to the focal point of the camera, then was rotated and moved to the position in the second frame. Let for a point of the object $(x_1, y_1)$ be its projection's coordinate in the first projection plane, and $(x_2, y_2)$ - in the second. $Z_1, Z_2$ be its position coordinates in space during the first and the second projection. $A$ be the rotation matrix. Then we get

$$x_2 Z_2 = (a_{11}x_1 + a_{12}y_1 + a_{13})Z_1 - a_{13}$$

$$y_2 Z_2 = (a_{21}x_1 + a_{22}y_1 + a_{23})Z_1 - a_{23}$$

$$Z_2 = (a_{31}x_1 + a_{32}y_1 + a_{33})Z_1 - a_{33}$$

This equation system contains two unknown dependent on the point $Z_1, Z_2$ that can be eliminated to yield a single equation containing only unknown parameters of rotation matrix $A$. We need 9 points in order to solve the equation system linearly for parameters of the rotation matrix $A$.

From the previous degrees-of-freedom consideration we know that the minimal number of traceable points is 5, but then the equation system gets highly non-linear [18]. It is worth mentioning, that the complexity of such a system lead Wang et al. [53] to the wrong conclusion, that four points and a line would be sufficient to recover the structure (non-linear equation system solving program "found" a solution though there is none).

If the traceable features are lines instead of points, three projections are needed instead of two [46, 60]. Linearization of the problem requires 13 lines [60].

## 3.3 Unknown and moving projection focal point

In [64] it has been demonstrated that two projections are insufficient for recovery of structure and motion under conditions of unknown and moving projection focal point. In [67] it has been suggested that 3 projections and 6 points would be sufficient. The epipolar geometry transformation of the original problem has been applied. But there may be some doubts about possibility of translating the epipolar geometry solution into the original geometry. The basic argument is that from 7 traceable points we can identify mutual positions of focal points between each pair of frames. Analyzing three pairs we find out that part of the information gained in this way is superfluous. Hence one may suspect that not all the information flows into solution of the problem.

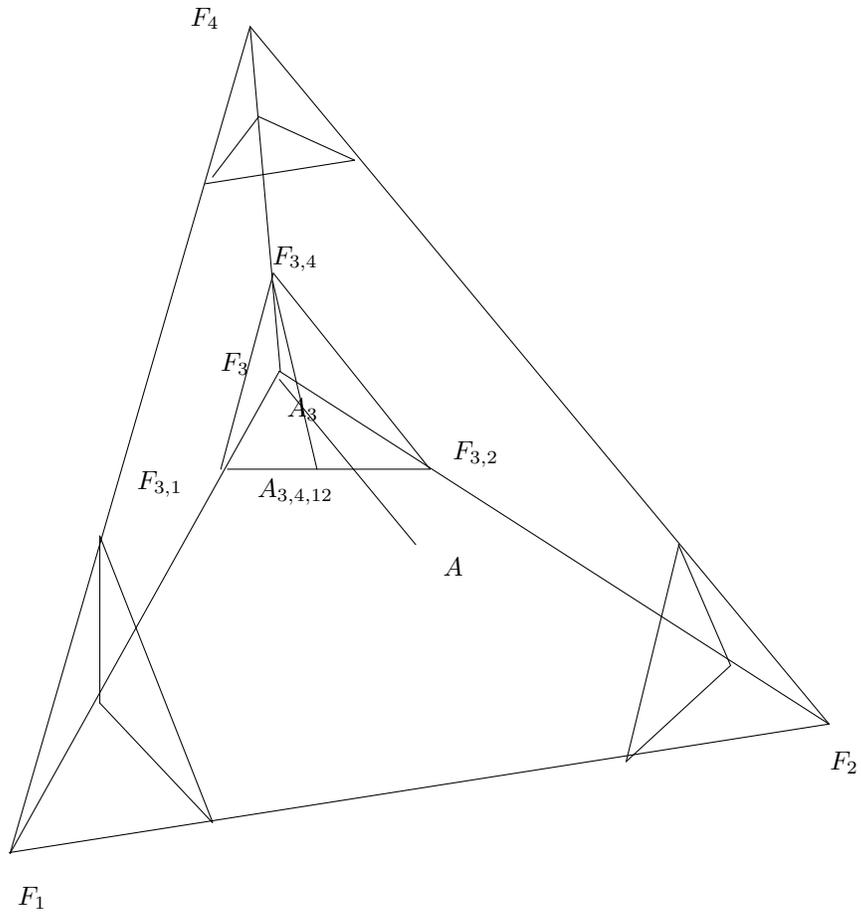

Figure 3: Four projection planes

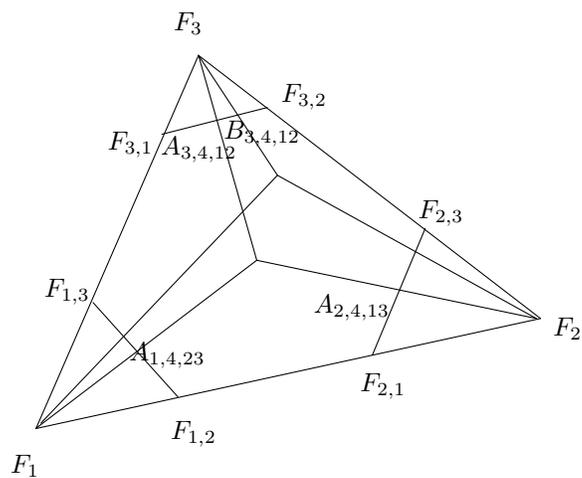

Figure 4: Four degrees of freedom in plane $F_1F_2F_3$.

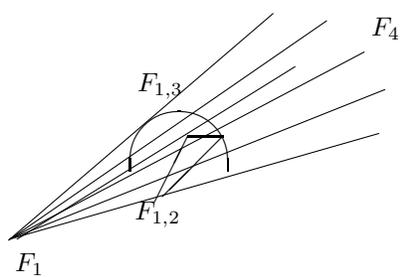

Figure 5: "Skewed cone" containing $F_4$.

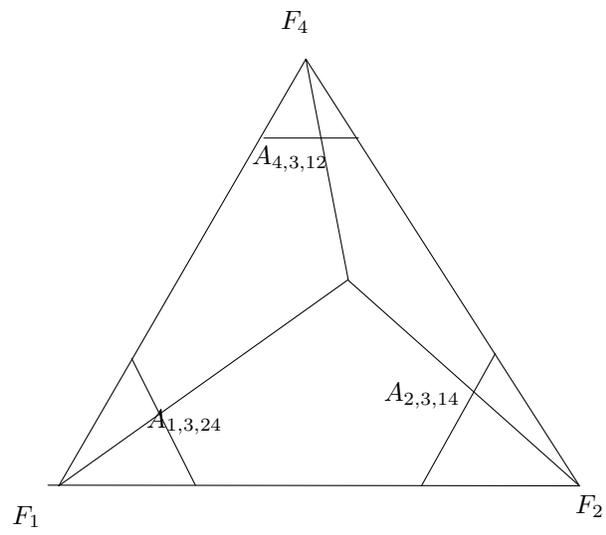

Figure 6: Three lines intersection in plane $F_1F_2F_4$.

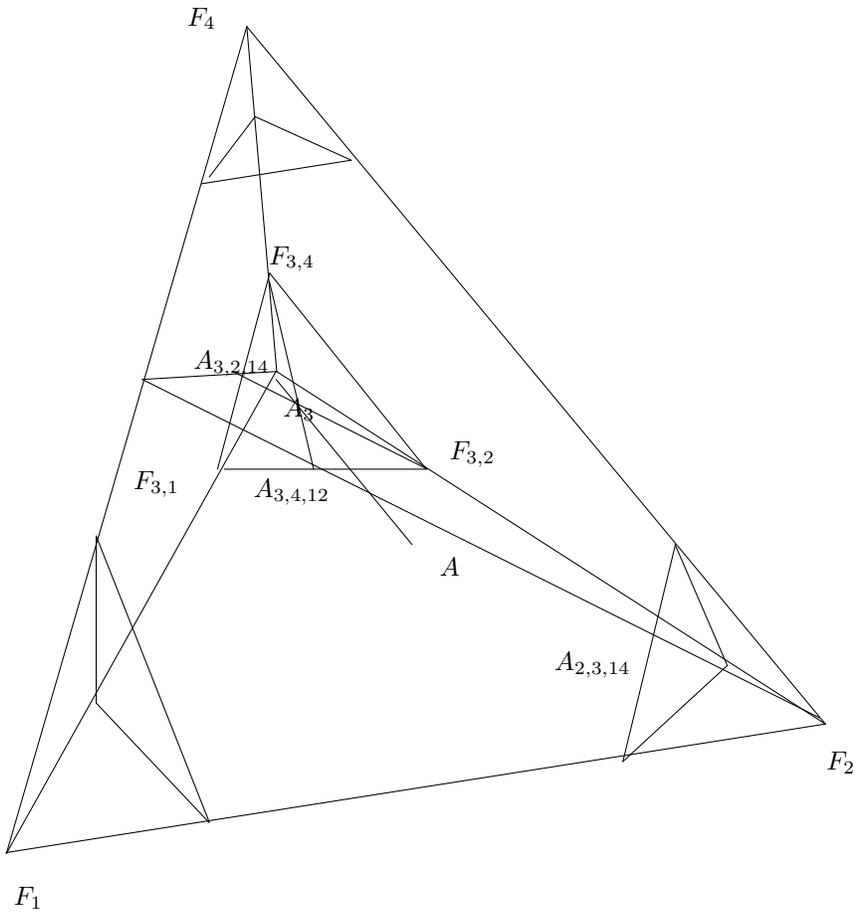

Figure 7: Line matching between neighbouring planes

Therefore we outline here recovery of structure and motion from four projections and seven traceable points. Instead of assuming the motion of the object, we assume the projection plane and focal point moving (Fig.3). Let $F_i$ denote the focal point of *ith* frame. Let $F_{i,j}$ be the projection of the *jth* frame focal point onto the *ith* frame. Let us consider only two object points $A, B$. Their projections onto the *ith* frame be denoted $A_i, B_i$. Let $A_{i,j,kl}$ denote the intersection of lines $F_{i,j}A_i$ and $F_{i,k}F_{i,l}$. Let us consider the plane $F_1, F_2, F_3$ (Fig.4). In this plane, distances between collinear points $F_{1,2}, F_{1,3}, A_{1,4,23}, B_{1,4,23}$ are known, also between collinear points $F_{2,1}, F_{2,3}, A_{2,4,13}, B_{2,4,13}$ and between collinear points $F_{3,1}, F_{3,2}, A_{3,4,13}, B_{3,4,12}$. If we assume the position of the points $F_1, F_2$ (we can do this because of undiscernibility of the scaling factor), due to distances between $F_{1,2}, F_{1,3}, A_{1,4,23}, B_{1,4,23}$ and between $F_{2,1}, F_{2,3}, A_{2,4,13}, B_{2,4,13}$ we have 4 degrees of freedom in selecting angles $F_3F_1F2, A_{1,4,23}F_1F2, F_3F_2F1, A_{2,4,13}F_2F1$. Taking into account the restriction of distances between $F_{3,1}, F_{3,2}, A_{3,4,13}, B_{3,4,12}$, we have practically only three degrees of freedom in selecting above-mentioned angles. Once we fixed them, we can derive the position of $F_3$, and of points $F_{1,2}, F_{1,3}, A_{1,4,23}, B_{1,4,23}, F_{2,1}, F_{2,3}, A_{2,4,13}, B_{2,4,13}, F_{3,1}, F_{3,2}, A_{3,4,13}, B_{3,4,12}$.

Having $F_1, F_{1,2}, F_{1,3}$ we can derive a "skewed cone" on which the point $F_4$ is lying (we know the distances $F_{1,2}F_{1,4}$ and $F_{1,3}F_{1,4}$) (Fig.5). Similarly with $F_2, F_{2,1}, F_{2,3}$ and $F_3, F_{3,1}, F_{3,2}$. So we can explicitly state the position of $F_4$ as the unique intersection point of these "cones".

In the plane $F_1, F_2, F_4$ we have no more freedom in selecting any points (Fig.6). We have to impose the condition that the three lines $F_1, A_{1,3,24}$, $F_2, A_{2,3,14}$, $F_4, A_{4,3,12}$ all intersect at the same single point. So we are left with two degrees of freedom. In the pair of planes $F_1, F_2, F_4$ and $F_1, F_3, F_4$ we have to require that the lines $F_2, A_{2,3,14}$, and $F_3, A_{3,2,14}$ intersect (Fig.7). In the pair of planes $F_1, F_2, F_4$ and $F_2, F_3, F_4$ we have to require that the lines $F_1, A_{1,3,24}$, and $F_3, A_{3,1,24}$ intersect. In this way we consume the remaining two degrees of freedom and get a final (highly non-linear) equation system to be solved in the angles $F_3F_1F2, A_{1,4,23}F_1F2, F_3F_2F1, A_{2,4,13}F_2F1$.

## 4   The problem of non-traceable points

Several authors devoted attention to recovery of the shape of curves, consisting essentially of non-traceable points e.g. [38] (under orthogonal projection), [66] (under epipolar geometry)

Let us assume we have already recovered the spatial positions of traceable points (the positions of cameras, projection planes). Let us assume that the nontraceable features are of the form of some smooth curves.

Let us consider projections of the object O on two known planes P' and P" with known focal points F' and F" respectively so that we know its both images O' and O". The principle of recovery of all non-traceable points (inner points of the smooth curve) is quite simple and goes as follows: First we select a non-traceable point, say D', of the first projection O', which is a projection of an unknown point D of the original curve, and recover the correspondence between D' and the unknown point D" being the projection of D onto the second plane P". We "draw" in our mind the straight line F'D' (its projection onto P' is just the point D') and project it then on the plane P" (using focal point F"). This projected line will surely cross the projected curve O" at (at least one) point D" which we claim to be the projection of the unknown point D (continuity resolves eventual ambiguities). In this way the so far non-traceable point D acquires the status of a traceable point (we know now its projections D' and D") and we can proceed with positioning the point D in space by "drawing" in space straight lines D'F' and D"F" and identify D as being the intersection point of both lines.

If orthogonal projection is considered, we shift focal points simply to infinity and proceed in the same way.